%% file: base.tex
\documentclass [twocolumn] {svjour3}

\smartqed  % flush right qed marks, e.g. at end of proof

\usepackage{url}
\usepackage{graphicx}
\usepackage[numbers, sort&compress]{natbib}
\usepackage{booktabs} % For formal tables
\usepackage[]{algorithm2e}
\usepackage{comment}
\usepackage[draft]{todonotes}
\makeatletter
\renewcommand\@makefntext[1]{\leftskip=1em\hskip-0.6em\@makefnmark#1}
\makeatother

\journalname{Künstliche Intelligenz}
\begin{document}

\title{I Feel I Feel You: A \emph{Theory of Mind} Experiment in Games}


%\subtitle{A \emph{Theory of Mind} Experiment in Games}


\author{David Melhart \and
        Georgios N. Yannakakis \and
        Antonios Liapis
}


\institute{David Melhart \at
\email{david.melhart@um.edu.mt}
\and
Georgios N. Yannakakis \at
\email{georgios.yannakakis@um.edu.mt}
\and
Antonios Liapis \at
\email{antonios.liapis@um.edu.mt}\\\\
Institute of Digital Games, 
University of Malta,
Msida, MSD 2080,
Malta\\
}

\date{Received: date / Accepted: date}

\maketitle

\input{body}

%\begin{acknowledgements}
%\end{acknowledgements}

\small
\bibliographystyle{spbasic}      % basic style, author-year citations
\bibliography{bibliography.bib}   % name your BibTeX data base

\end{document}

%% file: body.tex
\begin{abstract}
%219 words:
%How do we feel about the feelings of artificial agents we play against? Do we recognise their behaviour as emotional? Is it merely a cognitive process or does it involve observable affective manifestations as well? 

In this study into the player's emotional \emph{theory of mind} of gameplaying agents, we investigate how an agent's behaviour and the player's own performance and emotions shape the recognition of a frustrated behaviour. We focus on the perception of frustration as it is a prevalent affective experience in human-computer interaction. We present a testbed game tailored towards this end, in which a player competes against an agent with a frustration model based on theory. We collect gameplay data, an annotated ground truth about the player's appraisal of the agent's frustration, and apply face recognition to estimate the player's emotional state. We examine the collected data through correlation analysis and predictive machine learning models, and find that the player's observable emotions are not correlated highly with the perceived frustration of the agent. This suggests that our subject's \emph{theory of mind} is a cognitive process based on the gameplay context. Our predictive models---using ranking support vector machines---corroborate these results, yielding moderately accurate predictors of players' \emph{theory of mind}.

\keywords{Theory of Mind \and Affective Computing \and Digital Games \and Artificial Agents \and Preference Learning}

\end{abstract}

\section{Introduction \label{intro}}

%In 1997, Garry Kasparov was famously beaten by Deep Blue, the chess-playing computer of IBM. Recollections of the six games played attribute a great deal of Deep Blue's success to the $44^{th}$ move of the first game \cite{silver2012signal}. Deep Blue made an unexpected move due to a bug in its code, which Kasparov attributed to some superior form of intelligence. He admitted that he got anxious facing Deep Blue ever after. This example illustrates perfectly how attributing intelligence to artificial agents can lead to tangible emotional reactions. Kasparov's perspective on the behaviour of Deep Blue---even if internally it was a bug---is far from trivial for us to model. Arguably, it is generally complex to unravel how we feel other actors (humans or agents) feel. It is also largely unknown how we represent others' emotional and cognitive patterns according to the fundamental process known as the \emph{theory of mind} \cite{premack1978does,gopnik1992child,poletti2012cognitive}: the \emph{feeling of how others feel}.

Understanding how we recognise and feel about artificially simulated emotional behaviour is central to the design of believable characters featured in modern, narrative-heavy AAA games and the research of emotional modelling and affective computing. Arguably, it is generally complex to unravel how we feel other actors (humans or agents) feel. It is also largely unknown how we represent others' emotional and cognitive patterns according to the fundamental process known as the \emph{theory of mind} \cite{premack1978does,gopnik1992child,poletti2012cognitive}: the \emph{feeling of how others feel}.

Traditionally, the \emph{Theory of Mind} (ToM) refers to the mental models we form about others' higher order beliefs. However, recent studies shed light on the emotional components of ToM \cite{poletti2012cognitive,sebastian2011neural} as well. Throughout this paper we use a taxonomy of cognitive and emotional representation, which relies on the belief-order attribution hierarchy \cite{perner1985john}. According to this taxonomy we refer to our own beliefs and feelings as \emph{zero-order representation} and our mental model of another actor's beliefs and feelings as \emph{first-order representation}. In this regard, a \emph{second-order representation} would be another actor's recognition of our own judgement (i.e. ``it knows I know its state"). However, here we focus only on the players' recognition of emotion: specifically, their first-order representation of the agent's frustration.

We argue that modelling reliably a user's ToM can be viewed as the holy grail of not just user research and user experience design, but also adaptive and creative computation for any task that involves user-agent interactions. In games, modelling ToM could revolutionise adaptivity and personalisation---e.g. in the form of dynamic difficulty adjustment, procedural content generation, interactive narrative etc.---as our knowledge about the player's understanding of game agents would afford us more nuanced control over the experience \cite{yannakakis2018artificial}.

We explore the player's emotional ToM from both statistical and predictive modelling, investigating how the gameplay context and the player's emotional state during play affect their assessment of the agent's behaviour through two different lens. We process our metrics (both input and output) in an ordinal fashion, accounting for both absolute (i.e mean values) and relative (i.e. range of fluctuation) measures \cite{lopes2017ranktrace}. To conduct our experiments, we introduce the MAZING testbed game, in which the player competes against an artificial agent designed to exhibit frustrated behaviour based on a top-down model inspired by the theory of \emph{computer frustration} \cite{bessiere2006model}. We focus on frustration as one of the most prevalent and context-dependent affective outcomes of human-computer interaction. The main goals of our study are (a) to investigate the relationship between the gameplay context, manifestations of player emotion, and the first-order representation of the perceived frustration of the agent based on its behaviour, while (b) to explore different ways of processing the self-reported ToM.

This paper is novel to the field of games user research and affective computing as it introduces a player-centred approach to ToM in human-agent interaction. To the best of our knowledge, this is the first time ToM is examined within human-agent scenarios, where the focus is not on the model of the agent per se but rather on the players' first-order affective ToM process. Although most studies conceptualise ToM as a highly cognitive construct, we focus on the emotional component of the process and attempt to shed light on the ways players perceive how emotional agents feel.

\section{Theoretical Background}

This section provides the theoretical basis for our study and the agent model. We introduce the processes behind cognitive and affective aspects of ToM and present the theory of \emph{computer frustration} which inspired our top-down frustration model of the gameplaying agent.

\subsection{Theory of Mind and Emotions}
As briefly mentioned in Section \ref{intro}, the \emph{Theory of Mind} (ToM) is the concept of high-level mental models. Although traditional views focused on the representation of cognitive processes \cite{gopnik1992child}, the concept has been recently extended with an affective dimension \cite{poletti2012cognitive,sebastian2011neural}. ToM plays a central role in social cognition and interaction \cite{garfield2001social} as it enables humans to hold and manage prevalent representations of other actors, their beliefs, emotions, and cognitive processes.

ToM has been investigated from the late '70s \cite{premack1978does,bruner1981intention} and in the context of autonomous multiagent interaction from the mid-90s \cite{albrecht2017autonomous}. However, it is only recently being considered in game design and game user research. Although the bulk of studies focus on agent-based ToM modelling \cite{arrabales2009towards,de2015higher}, other venues consider \linebreak[4]player-player interactions \cite{hedden2002you,meijering2011know,goodie2012levels} and player-game involvement \cite{bormann2015immersed}. Motivated by the lack of a human-agent interaction perspective, this paper explores cognitive and emotional manifestations of a human's ToM while interacting with a game agent. While traditionally ToM is concerned with beliefs, trait judgements and strategic decisions \cite{schaafsma2015deconstructing}, we follow Damasio's \emph{somatic marker hypothesis} \cite{damasio1996somatic} and approach ToM from an emotion-centric perspective. 

Based on neuroscientific evidence, we differentiate between a cognitive and an affective ToM. \emph{Cognitive ToM} is focusing on belief and knowledge representations, while \emph{affective ToM} processes are involved in the representation of emotions \cite{shamay2010role}. However, these processes are not mutually exclusive \cite{sebastian2011neural}. Cognitive ToM is generally associated with brain regions involved in autonomic responses and a choice-selection downstream of the decision making process \cite{gallagher2003functional,weilbacher2016interplay}. 
Meanwhile, affective ToM involves additional areas tied to the affective and cognitive regulation of decision making processes as described in the {somatic marker hypothesis} \cite{damasio1996somatic,dunn2006somatic}. 
Evidence also shows that {affective ToM} relies on cognitive empathy, which is the understanding of others' emotions, and to a lesser degree on emotional contagion, a form of emotional mimicry \cite{sebastian2011neural}. This suggests that while it is possible to represent other actors' mental states cognitively, affective processes impact the formulation and regulation of such mental models.

The state of the art research in virtual agents, inferring goals and recognising false beliefs, is paving the way in developing bottom-up solutions for modelling artificial ToM \cite{rabinowitz2018machine}. Such approaches, however, generally do not consider modelling affective aspects of ToM \cite{arrabales2009towards,de2015higher,rabinowitz2018machine}. Adopting the typology of \cite{rabinowitz2018machine} to human players, we focus on agent-specific ToM---as opposed to general ToM, which stipulates a general predictive system---and turn our investigation towards how players formulate the cognitive and affective components of ToM with regards to game-playing agents.

\begin{figure}[!tb]
\centering
\includegraphics[width=0.8\columnwidth]{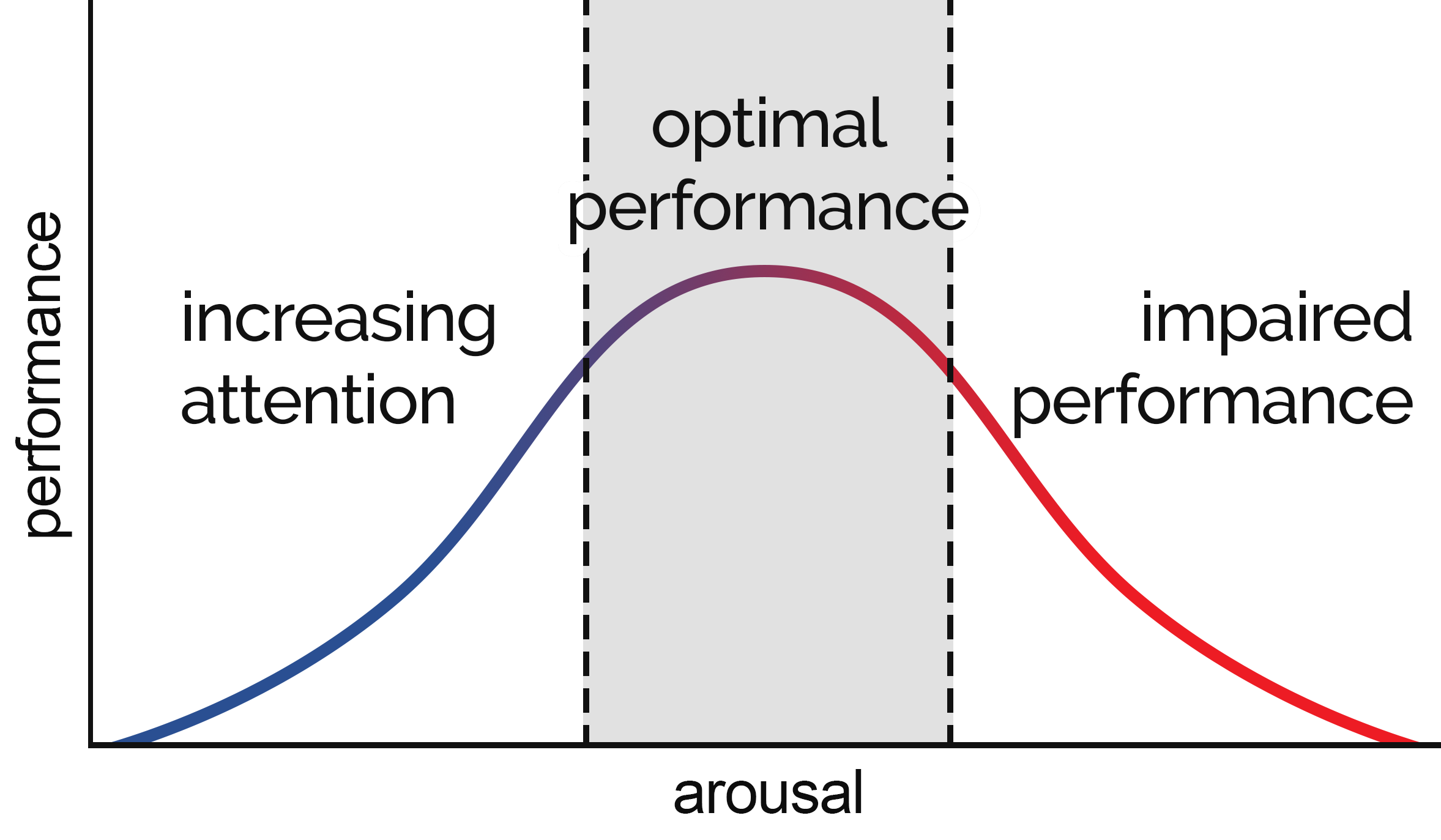}
\caption{Optimal level of human performance based on \cite{hebb1955drives}.} 
\label{fig:hebb}
\end{figure}

\subsection{Computer Frustration Theory}\label{sec:frustrationtheory}

This explorative first study of player-agent ToM addresses perceptions of frustration. Frustration is one of the most common complex affective responses experienced during human-computer interaction \cite{bessiere2006model}, with distinctive cognitive and behavioural patterns \cite{canossa2011arrrgghh}. The model we use for our game agents relies on the principles of the \emph{computer frustration} theory \cite{bessiere2006model} which is based on the work of \cite{hebb1955drives} and \cite{amsel1992frustration}. Computer frustration is a complex model which incorporates pre-emotional appraisal, immediate emotional response, and long lasting mood \cite{bessiere2006model}. Computer frustration is positioned within the information processing theories of cognition and emotion  %\cite{simon1967motivational}
\cite{ortony1990cognitive,rauterberg1995framework,carver2012attention}, by emphasising its role in pre-emotional appraisal.

According to the theory, frustration is triggered by the lack of anticipated change and manifests as \emph{non-specific arousal} in the information processing system, leading to an eventual cognitive performance dysfunction. {Computer frustration} differentiates between incident, session, and post-session frustration and focuses on self-efficacy, appraisal, and emotional outcomes of human-computer interaction. However, given the fast-paced nature of the game we designed for testing our hypotheses, in this paper we concentrate on the \emph{short-term} effects and functions of frustration. {Computer frustration} further predicts that the severity of the interruptions and the time lost are the primary causes of incident level (moment-to-moment) frustration---whereas low self-efficacy and negative mood have a greater effect on session level and post-session outcomes. 

However, not all frustrating events are detrimental to one's performance. Instead, {computer frustration} posits a bell-curve-like function between the level of arousal and performance \cite{bessiere2006model}, based on a Hebbian interpretation (Fig. \ref{fig:hebb}) of the {Yerkes-Dodson Law} \cite{yerkes1908relation}. Due to the connection between arousal and performance, frustration initially has a positive effect by limiting peripheral processes (both in perception and information processing), and thus helps focus on the task at hand. This enhancing effect is especially true if the frustration originates from unmet goals or expectations (\emph{in-game frustration}) rather than from a failure to operate an input device (\emph{at-game frustration}) \cite{gilleade2004using}.

\begin{figure}[!tb]
\centering
\includegraphics[width=1.0\columnwidth]{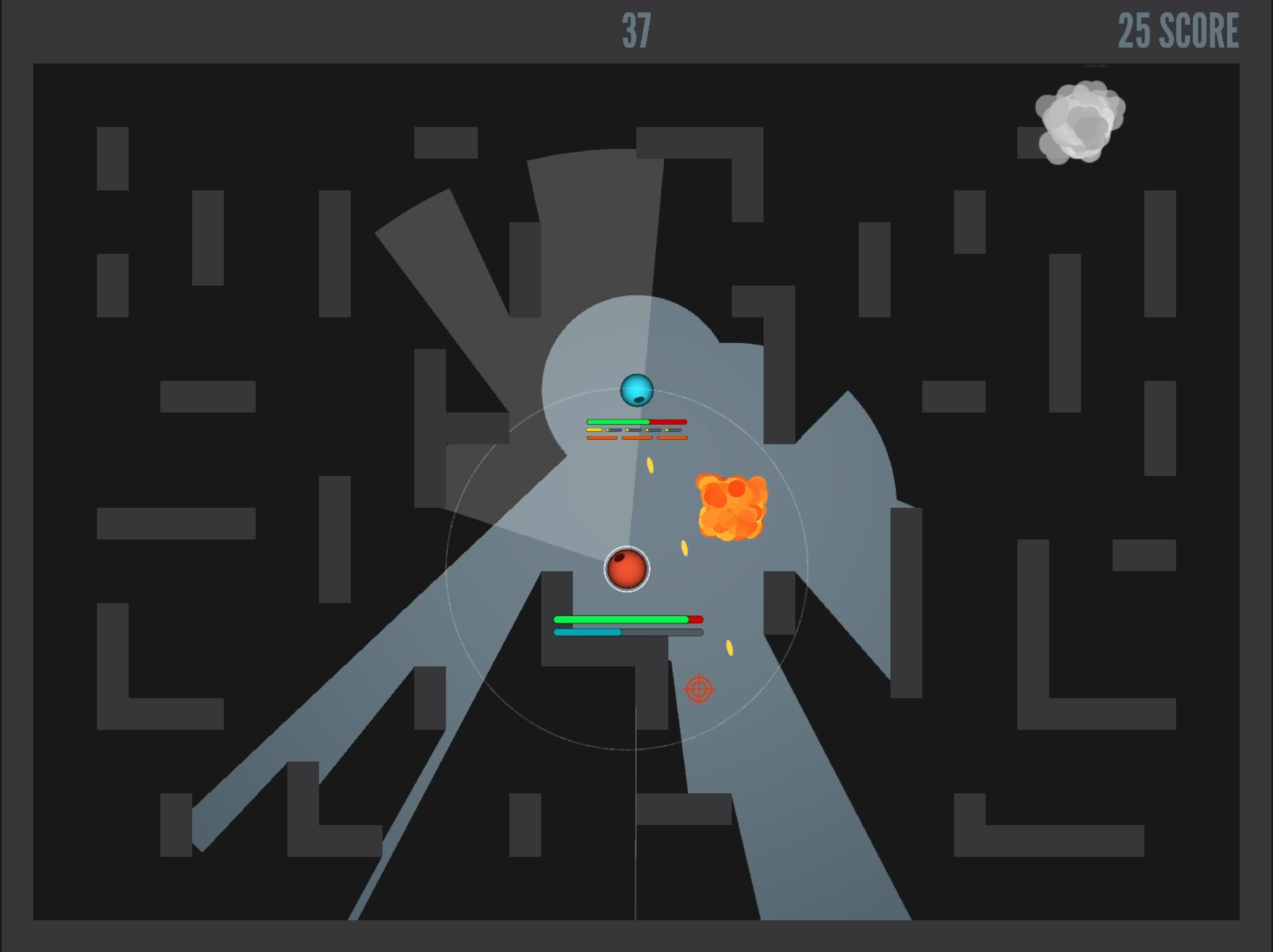}
\caption{Screenshot of \emph{MAZING}, showing the player attacking the agent (teal and red orbs) and a fire in the middle. A previously laid fire is disappearing in an upper corner.}
\label{fig:game}
\end{figure}

\section{The MAZING Game \label{game}}

To collect data on a game featuring an artificial agent that might exhibit frustration, we developed a 2D top-down shooter game where a player and an artificial agent compete (Fig. \ref{fig:game}). The player scores points by attacking the agent, while avoiding it. A game session automatically ends after 1 minute.

In this study, we collect data from 4 playthroughs per player: in each playthrough, the opponent is different in terms of its level of frustration. The first agent has no integrated model of frustration (the value of frustration remains at 0). The other three agents are reactive to their environment and vary their frustration scores according to our model between $25-50$, $50-75$, and $75-100$. In the following sections we detail the player's and agent's goals in this game.

\subsection{Player's Mechanics}

Players move in a 2D maze (viewed from a top-down perspective) using the \emph{WASD} keys and aiming with the mouse. Their movement speed is higher than the agent's base speed, giving the player an upper hand in most scenarios. They can also use a short \emph{dash} ability every 2 seconds, which grants them a speed multiplier for less than a second. The player scores points by damaging the agent, via two modes of attack: (a) shooting up to 5 projectiles in quick succession by holding down the left mouse-button, and (b) throwing bombs with the right mouse-button. Bombs create fires where they land for 5 seconds (see Fig. \ref{fig:game}). Passing through fire carpets deals damage to both the player and the agent, and agents are generally discouraged from moving through them. Both attacks recharge after a short period of time. If the agent dies, players gain additional points.
The game obscures the maze with a partial fog-of-war, which hinders visibility but does not block it completely. Players' avatars have their field of view which illuminates the map primarily in a cone in front of them and to a lesser extent peripherally (including behind the avatar) as shown in Fig.~\ref{fig:game}. Players lose if they collide with the agent or if they lose all their hit points (players lose hit points only when passing through fire carpets). Losing decreases the player's score and re-spawns the agent and the player at their original locations. 

\subsection{Agent's Mechanics} 

The agent only performs movement and low-level decision making. The agent carries out a basic search behaviour, quasi-randomly wandering around the map. At the end of each search cycle, it picks a random point and makes its way there, avoiding fires set by the player. If the agent senses the player, it engages in a chase. To sense the player, the agent possesses two distinct sensors mimicking visual and auditory senses. The visual sensor has an initial angle of 135$^{\circ}$ and a 10 metre radius. The auditory senses affect an area around the agent (initially also 10 metres), and have a low initial probability of detecting the player. If the player is standing within the sensor's reach, the agent's auditory system gradually increases the chance of detection and checks for the player every second. Intervening walls cut the auditory detection chance approximately in half. The agent takes damage from each bullet-hit and damage over time while standing in fire. The agent has many hit points, but they are not replenished over time.

\subsection{Agent's Frustration Model} \label{model} 

In order to provide the player with a quasi-believable, responsive agent, we create a model of frustration that drives the perception, movement and decision-making of the agent. 
Based on the theory of {computer frustration} (see Section \ref{sec:frustrationtheory}), we regard the severity of the setback as the primary variable for increasing frustration. As the agent's primary short-term goal is to catch the player, all incidents that make it harder for the agent to do so increase its frustration. These incidents include player attacks, increasing distance from the player, and losing sight of the player. Since we conceptualize frustration as a form of arousal, we also give a light increase to the agent's frustration value whenever it spots the player. Given that we wish to model incident level frustration, we gradually decrease the agent's frustration whenever it is engaged in search behaviour.

Several stimuli from the game environment affect the agent's level of frustration. Frustration increases if the path towards a goal calculated in the previous frame is shorter than the current path (which indicates new obstacles or a player getting away) and decreases (at a lower rate) if it is longer. Frustration is increased when the agent spots the player and when the agent loses sight of the player. Third, the agent's health has an effect on the agent's frustration: frustration increases with each projectile hit. Finally, frustration slowly decreases in ``resting periods", when the agent is in search behaviour. All modifiers to frustration are designed to provide players with more persistent feedback \cite{lankoski2007gameplay}, and ensure that the agent is getting more frustrated throughout the session and cannot easily revert to its baseline.

The agent manifests frustration in several perceptible ways:

\textbf{Sensory system:} Frustration causes increasingly focused attention by  decreasing the angle of the agent's field of view (FoV): a frustrated agent can  see further but at narrower angles, which can increase the chance to spot the player. Similarly, the area of the agent's auditory sensors is smaller as frustration rises, but the probability of hearing the player increases.

\textbf{Movement:} On a basic level, frustration increases the agent's movement speed and rotation speed linearly. This improves the agent's performance in spotting the player initially. However, at high frustration levels it produces erratic movements; coupled with the narrower field of view, this can result in lower accuracy. Frustration also decreases the number of turns in search behaviour, simulating increasingly agitated behaviour. 

\textbf{Decision Making:} 
Generally, the agent chooses more dangerous paths towards its goal when frustrated. The agent perceives paths through fire carpets as riskier; it is more likely to take a risky option the more health it has, or if a safe path to the player is considerably longer. Frustration affects the risk taking factor and biases the agent's behaviour towards being more reckless. 

\textbf{Behavioural Outcomes:} We designed our frustration model to reflect observations in \cite{canossa2011arrrgghh}. As frustration increased during play tests in that study, aggravated players took increasingly more and more risk, rushed forward, and paid less and less attention to their surroundings. In light of this research, we modify the agent's different systems to bias its behaviour towards this direction. The focused sensors, increased speed, and risk-taking behaviour is initially helpful for the agent, creating a \emph{focused} state and modelling the increased attention of the agent. As frustration rises, the system produces distinctly \emph{frustrated} behaviour, including hasty movements, reckless behaviour, and loss of peripheral senses. Higher frustration levels, however, lead to \emph{rage} signified by erratic, jerky movement and an almost complete shut down of the agent's sensors. This behaviour is a natural fallout of the model but it is also in line with the \emph{frustration-aggression hypothesis}~\cite{berkowitz1989frustration}.

\begin{figure}[tb]
\centering
\includegraphics[width=1\columnwidth]{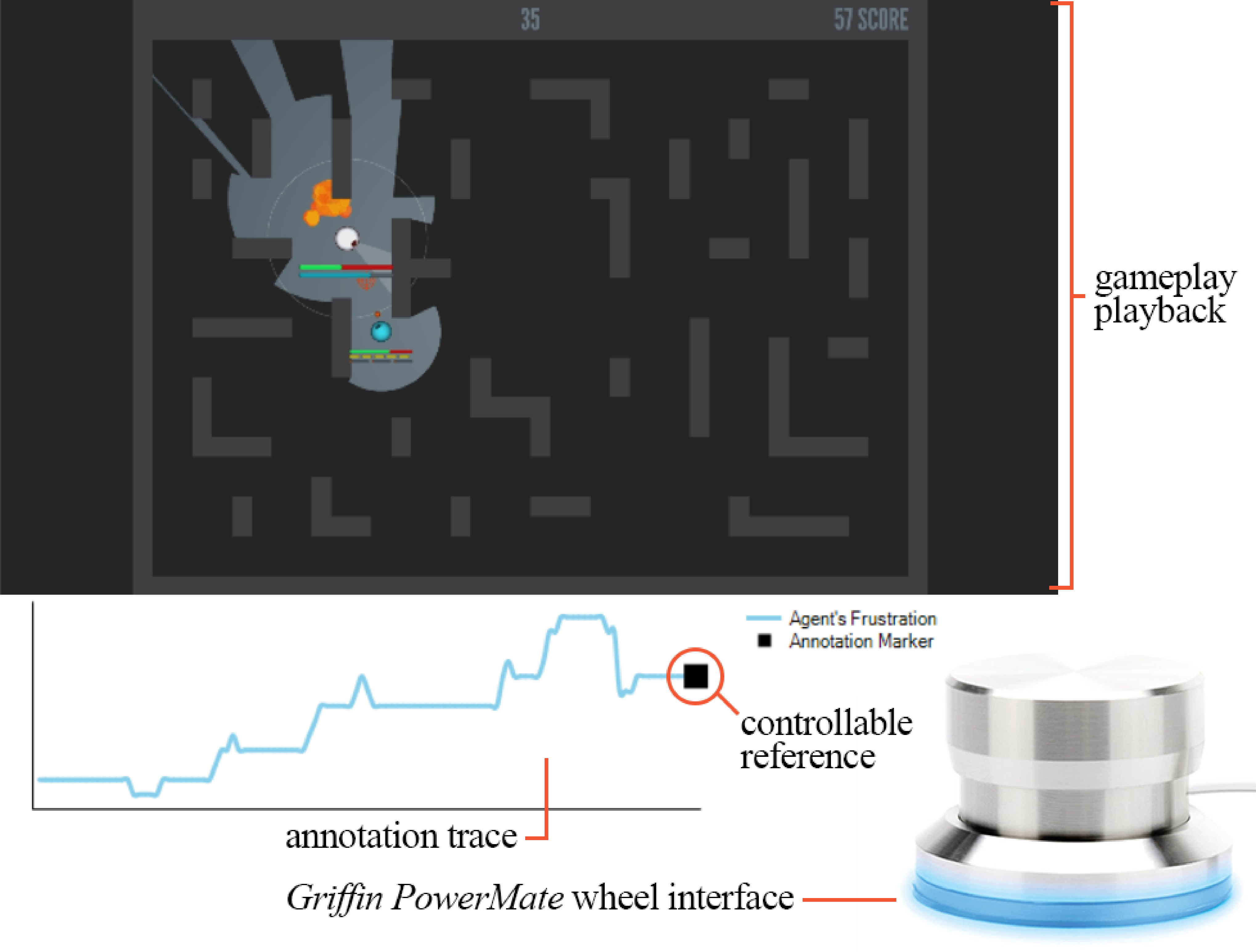}
\caption{The \emph{RankTrace} software annotation tool with its physical interface.} 
\label{fig:annotation}
\end{figure}

\section{Experimental Protocol\label{protocol}}

An experimental protocol was set up to collect data from each participant in a set of matchups with as diverse manifested frustration levels as possible. Each participant started with a tutorial level to get acquainted with the mechanics. After this, the participant played against an agent in 4 play sessions; each session was followed by a round of first-person annotation. During the 4 play sessions we recorded a number of gameplay metrics and players' facial features which were used to capture emotional manifestations during play. During the setup phase of the experiment, the facial recognition software was calibrated to each individual.

\subsection{Annotation \label{annotation}}

Following the core principles of ToM, we aim to assess what our players think about the feelings of the agents they have been observing and interacting with. Players were asked to annotate the first-order representation of the agent's frustration---i.e. \emph{how frustrated they felt the agent was}. To achieve this, the participant's last play session was recorded and played back to the player as a relived experience which the player annotated. 

Labels of the agent's frustration were collected via a continuous annotation process which offers a more reliable and detailed picture of the underlying ground truth \cite{yannakakis2015grounding} and captures the temporal dynamics of the experience \cite{mariooryad2013analysis}. Specifically, the players themselves annotated their perceived frustration of the agent in every game they played. Players used the \emph{RankTrace} tool (Fig.~\ref{fig:annotation}), which is an intuitive and validated \cite{lopes2017ranktrace,camilleri2017towards} annotation tool for unbounded and continuous annotation.  

The continuous frustration trace was then converted to ordered ranks between 3-second segments of gameplay. Processing trace annotation as ordinal data provides higher reliability, generality and inter-rater agreement \cite{yannakakis2015grounding,lopes2017ranktrace,camilleri2017towards} and is generally better aligned with the relative nature of emotions \cite{yannakakis2017ordinal}.

\subsection{Gameplay Features \label{gameplay-metrics}}

We extracted 30 features in each gameplay session which measure the position, kinaesthetic and sensory attributes and internal states of the agent. We also consider the position and actions of the player, and the interactions between the player and the agent (e.g. distance between player and agent). Collected features refer to (a) the agent's internal values: \emph{Frustration, Rotation Speed, Risk-Taking Factor, Movement Speed,  Hearing Radius, Hearing Probability, FoV Radius, FoV Angle, Number of Turns in Search}; (b) agent behaviour: \emph{Search Mode, Seeing Player, Chasing Player, Health, Distance Travelled, Taking Risky Path, Change in Rotation}; (c) player behaviour: \emph{Distance Travelled, Shooting, Pressing Shoot on Cool-down, Mouse Movement, Health, \linebreak[4]Dash Pressed, Dash Mode, Pressing Dash on Cool-down, \linebreak[4]Change in Rotation, Bomb Dropping, Pressing Bomb on Cool-down}; or (d) gameplay context: \emph{Score, Agent Distance From Player, Number of Fires}.

\subsection{Facial Emotion Recognition \label{facedata}}

Neuroscienctific evidence suggests that autonomic responses alone might not be sufficient when it comes to measuring ToM \cite{critchley2000cerebral,gallagher2003functional}. Emotional manifestations of ToM during gameplay are based on facial expression recognition and processing \cite{michel2003real}. We extract facial features and derive high-level facial expressions via the Affdex SDK \cite{mcduff2016affdex}. This system uses 34 facial landmarks to provide continuous feedback (with a rate of 10-30 FPS) and calculates the presence and intensity of the six basic emotions (anger, disgust, fear, joy, sadness, surprise) and contempt as well as estimates of the user's attention, engagement, and emotional valence from 14 facial action units. A total of 23 features are extracted from facial data captured during play and provided as intensity values of each expression on a scale between 0 (expressionless) to 100 (exaggerated display).

\section{Data Preprocessing and Methods \label{data}}

This section discusses our methods for data preprocessing and presents two quantitative measures of our signals, metrics and annotations. Subsection \ref{data_pl} offers a short introduction to preference learning, focusing on ranking Support Vector Machine (rankSVM) which is used to build predictive ToM models in Section \ref{results}.

\subsection{Data Format and Preprocessing \label{data_preproc}}
\begin{figure}[!tb]
\centering
\includegraphics[width=1.0\columnwidth]{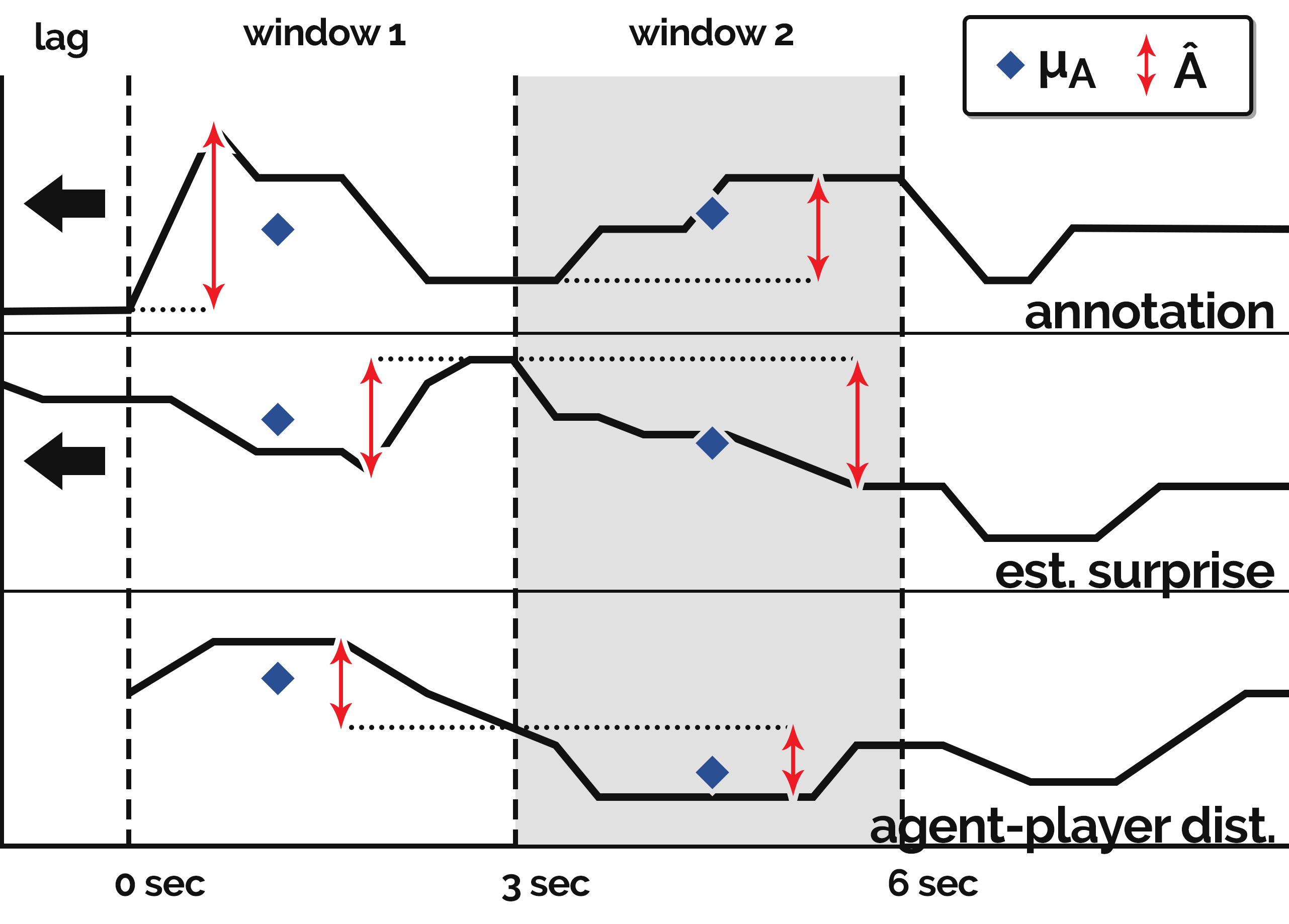}
\caption{
Calculating the mean and value range of different signals (top to bottom: player annotation, facial data, gameplay data) through a sliding window approach. Features are shifted back 1 second in relation to the gameplay metrics before cut into equal-length windows. Mean and value range are calculated from the highlighted time window (window 2) and its previous one (window 1) to derive rankings.} 
\label{fig:windows}
\end{figure}

Data from 80 play sessions is processed via a sliding-window approach. During this process the gameplay is segmented into consecutive equal-length windows with no overlap (\(w\)) and the mean value (\(\mu_{A}\)) and value range (\(\hat{A}\)) of each feature is calculated within each window (see Fig. \ref{fig:windows}). Both \(\mu_{A}\) and \(\hat{A}\) are relevant (and disparate): the mean values are an \emph{absolute metric} which is intuitive for comparing time windows (e.g.~whether the player believes the agent is more frustrated in one window than in the next). In contrast, value range measures the amount of change in the given metrics within a time window. While value range is expressed through absolute values as well, it captures the \emph{relative changes} within a time window. Ordinal relationships of value range between time windows can be intuitive for gameplay metrics or facial expressions (e.g.~whether there the game score changed more in one time window compared to the next) but, admittedly, are less intuitive for player annotation (e.g. whether the player saw a larger increase in agent frustration within one time window than in the next). Relative measures have been shown to be more powerful predictors than absolute ones for players' own affective states \cite{camilleri2017towards}. We believe that the degree of fluctuation within time windows can provide a clearer picture of the aggravated and erratic behaviour typical of frustrated players \cite{canossa2011arrrgghh} and the fluctuation of the player's appraisal of the agent.

Based on relevant findings \cite{mariooryad2013analysis}, we also consider the reaction lag of annotation traces and facial expressions (\(l\)). As in \cite{camilleri2017towards}, in this study we parse our data with a time window of 3 seconds (\(w=3\)), with no overlap between windows and a lag of 1 second (\(l=1\)). The lag is introduced to the annotation and facial features to account for the participants' reaction time. When calculating the lag, these values are shifted back (with the first 1 second discarded) before applying the windowing method. See Fig.~\ref{fig:windows} for an illustrative example.

\subsection{Method for Correlation Analysis \label{data_linear}}

We use Kendall's $\tau$ for all correlation analysis reported in Section \ref{results}. Kendall's $\tau$ is a non-parametric, bivariate test of correlation for measuring monotonic relationships \cite{nelsen2001kendall}, which is suited for analysing the concordance of ordinal data (unlike Pearson's correlation) and is a more robust metric than Spearman's $\rho$ but outputs lower correlation values \cite{fredricks2007relationship}. We treat significant findings at 5\% ($\alpha=0.05$) and highly significant at 1\% ($\alpha=0.01$) level. Because multiple comparisons are being made with the same variable (the processed annotation value) the Bonferroni correction is applied. Thus, the correlation analysis measures significance at $\alpha=\frac{0.05}{53}$ and high significance at $\alpha=\frac{0.01}{53}$ for each window-processing setup ($\mu_A$ and $\hat{A}$).

\subsection{Preference-based ToM Models \label{data_pl}}

Preference learning (PL) is a supervised learning technique, in which an algorithm predicts a rank order between two or more data points. The name \emph{preference learning} originates from the most prominent applications of these algorithms in predicting user preferences \cite{joachims2002optimizing}, however, as PL simply learns to predict ordinal relationships in the data, it can be used to solve a wide array of problems where it is important to conserve the relative relation of datapoints. We use PL to investigate the ordinal change in player's emotional ToM as there is a growing body of evidence that points towards the ordinal nature of emotions \cite{yannakakis2017ordinal,yannakakis2015ratings}, which underlines cognitive and affective processes. Even though we focus on ordinal changes, in this paper we use the term \emph{preference learning} to differentiate our algorithms from regression and classification algorithms. Contemporary research highlights the limitations of regression in affective computing \cite{yannakakis2018ordinal}. PL is also proving more robust than classification to handle ordinal annotations of affect \cite{camilleri2017towards,yannakakis2017ordinal,martinez2014don} as it preserves more information about the global and local order of the data than traditional class-based methods. 

In this study we use a form of pairwise preference learning, which leverages binary classification by transforming the representation of the dataset from singular datapoints to pairwise differences. During this transformation each pair of input points $(x_i,x_j) \in X^2$ are observed based on their  corresponding output $(y_i,y_j) \in Y^2$. Then a new dataset is constructed by assigning the pairwise difference of each pairs of input $x_i-x_j$ a label $\lambda=1$ and $x_j-x_i$ a label $\lambda=-1$ if $y_i>y_j$ (where $x_i$ is preferred over $x_j$). The resulting dataset reformulates the problem, which can be solved by any kind of binary classifier.

Because of the size of our dataset and the robustness of the technique, we use Support Vector Machines (SVM) for this task. SVMs are supervised learning algorithms, originally designed to solve classification problems by maximizing the margin of a separating boundary between data points \cite{vapnik1995statlearn}. Since their conception, SVMs have been adapted to solve different problems including regression analysis, clustering, and---in our case---ranking \cite{furnkranz2003pairwise}. % Ranking SVMs take advantage of one-against-one classification to learn the complete ranking of a dataset based on the preference relation of each provided example \cite{furnkranz2003pairwise}.
In our experiments, we use the SVM implementation found in the Preference Learning Toolbox\footnote{\url{plt.institutedigitalgames.com}} \cite{farrugia2015preference}, based on the algorithm of \cite{joachims2002optimizing}.

\section{Results \label{results}}
%(422 to 591, depending on the feature set) 
Following the experimental protocol of Section \ref{protocol}, we collected data from 20 participants (described in Section \ref{results_data}), processing them as ranks in terms of mean values and value range of subsequent time windows. These rankings are used to analyse the impact of individual features (with rank correlations presented in Section \ref{results_correlations}) and to train predictive models which combine some or all features linearly or non-linearly (in Section \ref{results_models}).

\subsection{Collected Data \label{results_data}}

Gameplay, facial and annotation data was collected from 20 participants (16 male, 4 female).
%10 participants (9 male, 1 female). 
Participants' average age was 30 %32, 
and all participants held or studied towards graduate degrees. All participants were experienced players, with half of them playing daily. % Still true in the new dataset /D/

Each participant played and annotated four gameplay sessions lasting 1 minute each. With a sliding window of 3 seconds, a total of $1,570$ data points are collected after partially missing data was removed. These errors were caused by limitations of the face detection software. To allow participants to play freely, a web-camera was used to record their faces. As some participants shifted in their chairs during gameplay, they inadvertently moved out from the camera's vision, resulting in missing facial data.
%800 windows of annotation and gameplay metrics are collected. For facial data, 661 windows are considered after unusable (erroneous or missing) data was removed due to limitations of the face detection software.

In Section \ref{results_correlations}, $1,570$ individual datapoints are considered, where each datapoint represents a 3 second snapshot of a player's gameplay.
%all $1,570$ data points are processed in terms of a global ranking to find correlations between annotations and individual features. 
For PL in Section \ref{results_models}, differences between all datapoints are considered. As discussed in Section \ref{data_pl}, for each comparison two observations are made and this results in $27,968$ comparisons for $\mu_A$ and $22,674$ comparisons for $\hat{A}$ with a 50\% baseline.

\subsection{Correlation Analysis \label{results_correlations}}

\begin{table}[!tb]
\caption{Kendall's $\tau$ correlation values between the annotation of frustration and features captured from the game and the web-cam. % of gameplay, player emotion, affect, and facial action units.
Values in bold are significant ($p<0.05$); highly significant values are underlined ($p<0.01$). Bonferroni correction is applied to all significance tests.
}
\centering
\begin{tabular}{l l r r}
\textbf{Type} & \textbf{Feature} &  \textbf{$\tau$}(\(\mu_{A}\))  &  \textbf{$\tau$}(\(\hat{A}\))  \\
\hline           
Agent model & Agent Frustration Score  &  0.048  &  0.038  \\
\hline           
Agent & Search Mode & \textbf{\underline{0.176}} &  -0.055  \\
behaviour & Seeing Player & \textbf{\underline{0.174}} & \textbf{\underline{0.134}} \\
 & Chasing Player & \textbf{\underline{0.169}} & \textbf{\underline{0.088}} \\
 & Distance Travelled & \textbf{\underline{0.125}} & \textbf{\underline{0.102}} \\
 & Rotation Speed & \textbf{\underline{0.101}} & \textbf{\underline{0.074}} \\
 & Speed  &  0.048  &  0.035  \\
 & Change in Rotation &  0.008  &  0.016  \\
 & Taking Risky Path  &  -0.002  &  0.008  \\
 & Search Mode Length &  -0.057  &  0.034  \\
\hline           
Agent & Health  & \textbf{\underline{-0.154}} & \textbf{0.065} \\
sensory & Hearing Probability &  0.054  &  0.033  \\
system & View Radius &  0.048  &  0.048  \\
 & Risk Taking Factor &  0.007  &  0.046  \\
 & Hearing Radius &  -0.049  &  0.040  \\
 & View Angle &  -0.049  &  0.035  \\
\hline           
Player & Shooting & \textbf{\underline{0.121}} &  0.044  \\
behaviour & Tries to Shoot on CD\footnotemark & \textbf{\underline{0.104}} & \textbf{0.073} \\
 & Distance Travelled & \textbf{\underline{0.072}} &  0.026  \\
 & Mouse Movement &  0.029  &  -0.028  \\
 & Change in Rotation &  0.029  &  0.040  \\
 & Health  &  0.018  &  0.033  \\
 & Tries to Bomb on CD &  0.014  &  0.047  \\
 & Dash Pressed &  -0.008  &  -0.047  \\
 & Dashing &  -0.009  &  -0.053  \\
 & Bomb Dropped &  -0.012  &  0.036  \\
 & Tries to Dash on CD &  -0.018  &  -0.027  \\
\hline           
General & Score  & \textbf{\underline{0.240}} & \textbf{0.070} \\
gameplay & Agent--Player Distance & \textbf{\underline{-0.141}} & \textbf{\underline{0.069}} \\
 & Number of Fires &  0.014  & \textbf{0.071} \\
\hline           
Basic & Contempt  &  0.037  &  -0.035  \\
emotions & Sadness  &  0.017  & \textbf{\underline{-0.071}} \\
 & Fear  &  0.009  &  -0.058  \\
 & Surprise  &  0.006  &  0.008  \\
 & Joy  &  0.002  &  0.019  \\
 & Anger  &  0.001  &  -0.041  \\
 & Disgust  &  -0.018  & \textbf{\underline{0.097}} \\
\hline           
Affective & Valence  & \textbf{\underline{0.077}} &  -0.044  \\
dimensions & Attention  &  -0.001  & \textbf{\underline{0.090}} \\
 & Engagement  & \textbf{\underline{-0.070}} &  0.000  \\
\hline           
Facial & ChinRaise  & \textbf{\underline{0.115}} &  0.040  \\
action units & BrowRaise  & \textbf{\underline{0.064}} & \textbf{0.066} \\
 & Smirk  &  0.028  &  -0.028  \\
 & InnerBrowRaise  &  0.027  & \textbf{\underline{-0.077}} \\
 & LipSuck  &  0.004  &  -0.017  \\
 & NoseWrinkle  &  -0.021  & \textbf{\underline{0.094}} \\
 & EyeClosure  &  -0.027  & \textbf{\underline{0.081}} \\
 & LipPucker  &  -0.029  &  0.025  \\
 & UpperLipRaise  &  -0.033  & \textbf{\underline{0.069}} \\
 & LipPress  &  -0.044  &  0.011  \\
 & BrowFurrow  & \textbf{-0.062} &  -0.053  \\
 & Smile  & \textbf{\underline{-0.067}} &  0.043  \\
 & MouthOpen  & \textbf{\underline{-0.100}} &  0.057  \\
\hline           
\end{tabular}
\label{table:corr}
\end{table}

Table \ref{table:corr} shows the Kendall's $\tau$ correlation values between annotated frustration of the agent and gameplay features of the agent, the player, and their interaction (i.e. General), as well as emotions estimated from facial detection.
%\footnote{Secondary captured facial features (e.g.~brow furrow) and less relevant play-metrics (e.g.~dash pressed on CD) are not shown in Table \ref{table:corr} as they lacked significant correlations.}
Correlations are calculated between the mean values (\(\mu_{A}\)) of features and the annotation data, and between the value range (\(\hat{A}\)) of a time window for features and the annotation data. As mentioned in Section \ref{data_linear}, we apply Bonferroni correction to all significance tests. Overall, there are only a handful of significant correlations in both \(\mu_{A}\) (18 out of 53 with $p<0.05$) and \(\hat{A}\) (17 out of 53 with $p<0.05$) cases. While most of the action units and more complex emotional and affective constructs measured by face recognition show very weak correlations (generally below 0.1), features relating to the agent's behaviour and the gameplay context show much stronger connections with the perceived frustration of the agent. 

\footnotetext{CD: Cool-Down. An ability is recharging and unavailable.}
%It is evident that there is a large amount of significant---albeit mostly weak---effects between the perceived frustration and the agent's internal frustration value, the agent-specific behavioural features (6 out of 8 with $p<0.1\%$), the agent's sensory system (5 out of 5 with $p<0.1\%$), some player-specific behavioural features (2 out of 7) and general metrics when the data is compartmentalised as $\mu_{A}$. %It is evident that there are strong effects between perceived frustration and most of the agent-specific behavioural features evaluated (5 out of 7 with $p<0.1\%$), but also player-specific behavioural features (2 out of 7) and general metrics. 
Perhaps surprisingly, the absolute highest correlation is with the player's score---which naturally relies both on the player's and the agent's performance. Even though other gameplay features inform the score (i.e. the agent's health), score is the utmost indicator of the success and failure of the player. Therefore, it provides additional high-level information about the game state compared to other, simpler features.
It is also evident that captured facial features including expressions of the six basic emotions \cite{ekman1980facial} and \emph{contempt} show even weaker correlations when the data is processed as $\hat{\mu}$ and no significant connections when it is processed as $\mu_{A}$.
%were not correlated with annotated frustration (with one exception that does not hold for higher significance thresholds). 
Since annotations of agent frustration have few significant correlations with affective markers but many significant correlations with contextual gameplay information, we may conclude that the first-order representation of the agents in our experiments is a predominantly cognitive process. 
%The lack of significant correlations between affective markers and annotation of agent frustration and the high number of significant correlations between contextual gameplay information and the perceived agent frustration could suggest that the first-order representation of the agents in our experiments is a predominantly cognitive process. 

\subsection{Predictive Models \label{results_models}}

While a traditional correlation analysis can indicate which individual features are good predictors of player ToM, it does not test how these features perform when combined in linear or non-linear fashions. We use preference learning methods (see Section \ref{data_pl}) to construct models based on different feature sets and with the input and output processed either in an absolute or a relative fashion. The input features consist of 30 gameplay features, 23 facial emotion manifestation features, and their combination (see these 53 features in Section \ref{gameplay-metrics} and on Table \ref{table:corr}). The output of our models are the ordinal relation of pairwise differences between datapoints. We infer these relation both in terms of mean values ($\mu_A$) and value ranges ($\hat{A}$). We indicate the processing of the input and output features with a right arrow between them (i.e. in case of an input processed as mean values and output processed as value ranges the notation is $\mu_A\rightarrow\hat{A}$). To test the robustness of our models, we apply cross-participant validation: i.e. training the model on data of 18 %9 
players and testing it on data of two unseen players, repeated 10 times so that all players are validated. To measure the statistical significance of the difference between models, two-tailed t-tests are used with $p<0.05$. When a model is tested against multiple other models, the calculation of significance is adjusted using the Bonferroni correction. %The correction is applied when comparing the performance of one model to the other models. 
Since 12 different models are compared, one model is significantly different from all others at $\alpha=\frac{0.05}{11}$.

\begin{figure}[!tb]
\centering
\includegraphics[width=1.0\columnwidth]{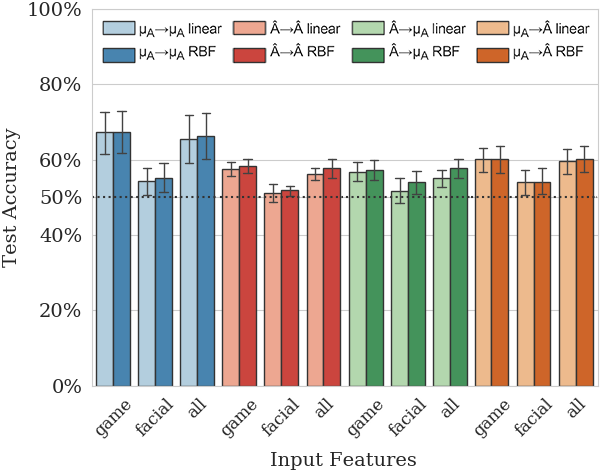}
\caption{Accuracies of linear SVM and best RBF SVM predictive models, on different combinations of \(\mu_A\) and \(\hat{A}\) input and output values ($input\rightarrow output$). Results are averaged from 10-fold %leave-one-participant-out 
cross-validation folds, and error bars denote the 95\% confidence intervals.}
\label{fig:predictive_models}
\end{figure}

Figure \ref{fig:predictive_models} shows the 10-fold %leave-one-participant-out
cross-validation accuracies of linear support vector machines (SVMs) and non-linear SVMs with Radial Basis Function (RBF) kernels. RBF emphasises the local proximity between input vectors in a feature space, allowing for a non-linear measure of match between vectors \cite{vapnik1995statlearn}. Both linear and RBF SVMs use the $C$ regularisation parameter to optimise the trade-off between maximising the separating margin and minimising the classification error, while the RBF SVMs also rely on the $\gamma$ hyperparameter to control the weight given to datapoints during the kernel calculation. The input features and output (annotations) are processed as mean value and value range separately, leading to four combinations of input--output. Results shown in Fig.~\ref{fig:predictive_models} are from the best $C$ and RBF$\gamma$ values per model\footnote{The best ($C$) and RBF$\gamma$ values for each model are:\newline
$\mu_A \rightarrow \mu_A$: game: (0.1) 0.5; facial: (100) 0.1; all: (1) 0.01.\newline
$\hat{A} \rightarrow \hat{A}$: game: (0.1) 0.5; facial: (10) 1; all: (0.1) 0.1.\newline
$\hat{A} \rightarrow  \mu_A$: game: (0.1) 0.1; facial: (0.1) 0.01; all: (0.1) 0.5.\newline
$\mu_A \rightarrow \hat{A}$: game: (0.5) 0.01; facial: (0.1) 0.01; all: (0.5) 0.01.
}
based on an exhaustive gradient search for both the $C$ and $RBF\gamma$ parameters from $10^{-3}$ to $10^3$ with powers of 10.

From Figure \ref{fig:predictive_models} it is evident that the modelling of perceived frustration is a challenging predictive task. Both linear and non-linear SVMs are performing less than 10\% above the baseline, with the exception of $\mu_A \rightarrow \mu_A$, which reaches 67.5\% on average (80.2\% at best) based on game features and 66.4\% on average (81.7\% at best) based on all features. Based on the results of the correlation analysis presented on Table \ref{table:corr}, it is not surprising that features processed as $\hat{A}$ yield weaker results: the best $\hat{A} \rightarrow \hat{A}$ models only reaching 58.4\% on average and 62.9\% at best based on gameplay features. The best model combining both processing techniques is $\mu_A \rightarrow \hat{A}$ with 60.2\% average and 69\% maximum accuracy using gameplay features. While the best models are achieved using gameplay features only, accuracies for models based on facial features corroborate findings of Section \ref{results_correlations}. In the $\mu_A \rightarrow \mu_A$ and $\hat{A} \rightarrow \hat{A}$ scenarios, models based on facial recognition result in significantly worse accuracies than other feature sets. There is no significant difference, however, between models using only gameplay features and when both feature sets are combined together in a bimodal fashion. The same significant differences are found with Mann–Whitney tests, which does not assume a normal distribution, further supporting our conclusions.

%The results of these predictive models are supported by Table \ref{table:corr}, which showed significant but weak correlations between perceived frustration and a number of gameplay features, and weak to no correlations between perceived frustration and facial features.
These results are supported by Table \ref{table:corr} and align with the conclusion of Section \ref{results_correlations} suggesting that the first-order representation of the agent mainly relies on the cognitive understanding of observable information without much emotional feedback. The deliberative nature of this mechanism might explain the weak predictions based on non-linear models, as players might actively interpret and reflect on the context of the interaction instead of relying on their own affective response or simple observations of the game state.

\section{Discussion \label{discussion}}

%a) what did we do?
This study examined the player's theory of mind regarding a gameplaying artificial agent which was designed to exhibit behavioural signs of frustration. The test-bed game, MAZING, was designed for the study based a contemporary theory of frustration in human-computer interaction. Within MAZING, an AI opponent was designed for the player to interact with. We collected first-person annotations of the player's first-order representation of the agent's frustration and examined the player's perception both through correlation analysis and via predictive models.

%b) what is the conclusion and contribution?
Results indicate that the most prominent correlations of the player's appraisal of agent frustration is the gameplay context, i.e. the performance and interaction of both player and agent. Our results also suggest that the process of developing and maintaining a ToM primarily relies on the understanding of the gameplay context, %cognitive process, 
with no strong monotonic correlations to visible signs of player emotion. Predictive models of a player's ToM showed that gameplay features alone are more reliable predictors of how players appraise situations and perceive agent behaviour and frustration. On the other hand, SVM models only had moderate success in predicting players' ToM. 
%The similar performance of linear and non-linear models suggests that further improvements of the modelling is hard to achieve through increasing modelling complexity and rather point towards latent features that could describe the underlying cognitive process better.

Our results are corroborated by \cite{fernandez2013advances} and recent findings of \cite{roohi2018facial}, which applied deep neural networks to the mapping of basic emotions to gameplay events with mixed outcomes. Just as their results, our research also indicates that the ambiguity and underlying complexity of emotions are not trivial to read and contextualise through facial emotion manifestations, leading to inaccurate predictions based on absolute measures of basic emotions. The meta-review of \cite{poria2017review} found that multimodal modelling generally outperforms unimodal methods in audio-, video-, and text-based analysis. Our results expand these findings to new modalities which we capture through the gameplay logs (i.e. player behaviour and gameplay context) and provide additional validity by showcasing the improved performance of models using both gameplay-based, and video-based (facial features) modalities.

%c) limitations & future work
The primary limitation of our study is the ad-hoc nature of the agent's model of frustration: while the model is inspired by contemporary theory and manifests a varied but persistent behaviour, the testbed cannot be validated based on the statistical analysis. Preliminary comparisons between players' annotations did not show substantial differences between playthroughs with agents exhibiting low or high frustration, but future work should find a more granular method of validating the internal models of the gameplaying agent through experimentation. This could involve a focus on basic, more universally recognised emotions, a more expressive agent, and more streamlined gameplay. 

Another limitation was the lack of a ground truth for the player's own emotional state, as we relied instead on detected emotion via facial expressions. While the correlation analysis showed little relationship between player emotion and perceptions of frustration, this could be instead due to the instrument used to capture emotion in the first place. We deliberately avoided an extra step asking players to annotate their own emotion, as this would cause more cognitive load and bias the ToM annotation due to ordering effects. However, future work could explore ways of collecting ground truth data on the emotional state of the players without increasing the difficulty of the annotation task. 

Finally, while this first study focused only on gameplay metrics and facial features, future work could extend the data collection to other modalites. This study also collected a number of physiological signals (heart rate variability and electrodermal activity), but due to varying signal quality we chose to omit them from this paper. Improved ways of collecting physiological signals, gaze tracking, or other ways to process the features in a relative fashion such as average gradient per time window \cite{camilleri2017towards} could lead to more robust predictive models, and should be further investigated.

\section{Conclusions \label{conclusions}}

This paper examined a player's theory of mind regarding an agent's simulated frustration. The MAZING test-bed game was created explicitly towards this end, inspired by the theory of computer frustration. Results from a small-scale study with 20 %10 
players gave us a rich dataset of granular annotations of perceived agent frustration, as well as 53 %54
features of gameplay and players' facial expressions. The analysis of the results indicated that a player's first-order representation of the agent's state is largely a cognitive process. Further, emotional responses were deemed unreliable in modelling player ToM, as relying solely on gameplay features yields models of significantly higher accuracies compared to models based on facial features.% and marginally higher accuracies compared to models built on the combined feature set.
%while emotions were not captured in linear models when combined non-linearly with the context (i.e. gameplay) they seem to have a strong predictive impact on modelling player ToM, especially regarding how fluctuations of perceived frustration can vary over time.